%% file: main.tex
\documentclass{article}
\pdfoutput=1
\usepackage{spconf,amsmath,graphicx}

\usepackage[unicode]{hyperref}

\usepackage{amsmath}
\usepackage{cleveref}
\usepackage{amssymb}
\usepackage{amsthm}
\usepackage{mathrsfs} 
\usepackage{textgreek}
\usepackage{bbm}
\usepackage{breqn}
\usepackage{multirow}
\usepackage{listings}
\usepackage{xcolor}
\usepackage{tablefootnote}
\usepackage{lipsum}

\newcommand\blfootnote[1]{%
  \begingroup
  \renewcommand\thefootnote{}\footnote{#1}%
  \addtocounter{footnote}{-1}%
  \endgroup
}

\usepackage{mathtools}
\usepackage{cuted}
\definecolor{codegreen}{rgb}{0,0.6,0}
\definecolor{codegray}{rgb}{0.5,0.5,0.5}
\definecolor{codepurple}{rgb}{0.58,0,0.82}
\definecolor{backcolour}{rgb}{0.95,0.95,0.92}

\lstdefinestyle{mystyle}{
    backgroundcolor=\color{backcolour},   
    commentstyle=\color{codegreen},
    keywordstyle=\color{magenta},
    numberstyle=\tiny\color{codegray},
    stringstyle=\color{codepurple},
    basicstyle=\ttfamily\footnotesize,
    breakatwhitespace=false,         
    breaklines=true,                 
    captionpos=b,                    
    keepspaces=true,                 
    numbers=left,                    
    numbersep=5pt,                  
    showspaces=false,                
    showstringspaces=false,
    showtabs=false,                  
    tabsize=2
}

\input{math_commands.tex}

\title{Role of Bias Terms in Dot-Product Attention}
\name{Mahdi Namazifar, Devamanyu Hazarika, Dilek Hakkani-T\"ur}
\address{Amazon Alexa AI}
\begin{document}
%\ninept
%
\maketitle
\begin{abstract}
Dot-product attention is a core module in the present generation of neural network models, particularly transformers, and is being leveraged across numerous areas such as natural language processing and computer vision. This attention module is comprised of three linear transformations, namely \textit{query}, \textit{key}, and \textit{value} linear transformations, each of which has a bias term. In this work, we study the role of these bias terms, and mathematically show that the bias term of the \textit{key} linear transformation is redundant and could be omitted without any impact on the attention module. Moreover, we argue that the bias term of the \textit{value} linear transformation has a more prominent role than that of the bias term of the \textit{query} linear transformation. We empirically verify these findings through multiple experiments on language modeling, natural language understanding, and natural language generation tasks.
\end{abstract}
\begin{keywords}
Attention, Transformer, Softmax\blfootnote{© 2023 IEEE. Personal use of this material is permitted. Permission from IEEE must be obtained for all other uses, in any current or future media, including reprinting/republishing this material for advertising or promotional purposes, creating new collective works, for resale or redistribution to servers or lists, or reuse of any copyrighted component of this work in other works.}
\end{keywords}

\section{Introduction}
Attention mechanism has revolutionized the application of neural networks in numerous areas such as computer vision and natural language processing. Different mechanism of attention such as Additive Attention \cite{38ed090f8de94fb3b0b46b86f9133623}, Multiplicative Attention \cite{luong-etal-2015-effective}, and Key-Value Attention \cite{DBLP:journals/corr/DanilukRWR17} have been introduced in the past. Among all different attention mechanisms, perhaps the one that is used most frequently is the Dot-Product Attention \cite{NIPS2017_3f5ee243} that was introduced for transformers.  Hereon, any mention of attention refers to this Dot-Product attention. 

% The current generation of pre-trained transformer-based models share a common lineage that started from~\cite{DBLP:conf/nips/VaswaniSPUJGKP17}. 
In the recent past, the attention module has been analyzed by various works, primarily in attempts to improve its squared complexity and aid its feasibility for long sequences~\cite{tay2020efficient}. Additionally, other works have analyzed potential redundancies in components like the multiple attention heads~\cite{michel2019sixteen,behnke-heafield-2020-losing,voita2019analyzing}.
%or the placement of modules like the location of layer norms~\cite{DBLP:conf/icml/XiongYHZZXZLWL20}. 
While these variants have shown promising evidence, the original transformer seems to be performing the best when compared in various scales of model sizes~\cite{tay2022scaling}. In this work, we analyze the attention module and attempt to dissect it to better understand its components.

Attention mechanism includes three linear transformations, namely \textit{query}, \textit{key}, and \textit{value} transformations, which are affine transformations with respective bias terms. In this work, we study the role of these bias terms, and mathematically show that the bias term for the \textit{key} linear transformation does not have any role in the attention function and could be omitted altogether. This result was also independently reported in \cite{DBLP:journals/corr/abs-2006-16362}. We next verify this result numerically, and show that replacing these bias vectors with arbitrary and random vectors does not result in any significant difference\footnote{The difference is non-zero due to numerical errors. See \Cref{sec:change_b_k} for details.} in the output of transformers. 
Another implication of this result is in BitFit \cite{ben-zaken-etal-2022-bitfit} where only bias terms in a transformer-based language model are fine-tuned on downstream tasks for parameter-efficient training. We show that by freezing the \textit{key} bias parameters in attention, we could reduce the number of trainable parameters in BitFit by over 11\% with no impact on the performance of the final model on downstream tasks.

\section{Notation}
\label{sec:notation}
In attention mechanism, an input vector $\vh \in \mathbb{R}^{d}$, attends to a set of $n$ vectors (subject of attention) which are represented as columns of matrix $\mC \in \mathbb{R}^{d\times n}$.
Within the attention mechanism, first a query vector $\vq \in \mathbb{R}^{d}$ is constructed based on $\vh$ using a linear transformation, i.e., $\vq =  \mW_q \vh + \vb_q$, where $\mW_q \in \mathbb{R}^{d \times d}$ and $\vb_q \in \mathbb{R}^{d}$ are the respective weight and bias parameters. Also, a set of key vectors that establish a key matrix $\mK \in \mathbb{R}^{d\times n}$ are constructed by applying another linear transformation on $\mC$, i.e., $\mK = \mW_k\mC+ \vb_k \mathbbm{1}^T$, where $\mathbbm{1} \in \mathbb{R}^{n}$ is a vector of 1s. Next, score distribution between the query and the keys is created by applying softmax ($\sigma(.)$) on the product of these two linear transformations:
\begin{align*}
&\sigma\left((\mW_q\vh+\vb_q)^T\left(\mW_k\mC+ \vb_k\mathbbm{1}^T)\right)\right) =  \sigma\left(\vq ^T\mK\right),
\end{align*}

which is referred to as attention distribution. On the other hand, similar to the process of creating the set of key vectors ($\mK$), a set of value vectors, constituting matrix $\mV$, are created by applying another linear transformation on $\mC$, i.e., $\mV = (\mW_v\mC+ \vb_v\mathbbm{1}^T)$. Finally, attention is computed by multiplying the value vectors and the attention distribution:
\begin{align}
\label{eq:attn_simple}
\text{Attn}&^{\mC}(\vh)= \mV\sigma\left(\vq ^T \mK\right)^T,
\end{align}

% &\sigma\left(( \vh \mW_q+ \vb_q)\left( \mC \mW_k+\mathbbm{1}^T \vb_k\right)^T\right)\cdot\\
% &( \mC \mW_v+ \vb_v\mathbbm{1}^T)^T
% where
% \begin{align}
%      \vq &=  \mW_q \vh+ \vb_q \label{eq:q}\\
%      \mK &=  \mW_k \mC+ \vb_k\mathbbm{1}^T \label{eq:K}\\
%      \mV &=  \mW_v \mC+ \vb_v\mathbbm{1}^T
% \end{align}
which could be thought of as a convex combination of value vectors (columns of $\mV$).\footnote{Note that we omit writing the scaling factor $\frac{1}{\sqrt{d}}$ employed on the dot-product, for brevity.}

\section{Dissecting Attention}
\label{sec:attn_dissec}
To better understand the inter-workings of attention, we take a deeper look into the interactions of the different elements that form the attention mechanism. First, we expand $\mV$ in \Cref{eq:attn_simple}:
\begin{align}
    &\text{Attn}^{ \mC}( \vh) = \mV\sigma\left( \vq ^T \mK\right)^T   \nonumber\\
    &\quad =( \mW_v \vh+ \vb_v\mathbbm{1}^T)\sigma\left( \vq ^T \mK\right)^T \nonumber\\
    &\quad =\mW_v \vh \sigma\left( \vq ^T \mK\right)^T + \vb_v\mathbbm{1}^T\sigma\left( \vq ^T \mK\right)^T.
    \label{eq:attn-open-1}
\end{align}

Since $ \vb_v\mathbbm{1}^T$ is a matrix with identical columns $\vb_v$, any convex combination of its columns, including the one using $\sigma\left( \vq ^T \mK\right)$ as weights, is essentially equal to $\vb_v$, i.e.,
\begin{align*}
    \vb_v\mathbbm{1}^T\sigma\left( \vq ^T \mK\right)^T  =  \vb_v.
\end{align*}
Therefore, from \Cref{eq:attn-open-1}, we can write the attention function as,
\begin{align}
     \text{Attn}&^{ \mC}( \vh)=\mW_v \vh\sigma\left( \vq ^T \mK\right)^T  +  \vb_v .
     \label{eq:attn-open-V}
\end{align}
Next, we expand $ \mK$ in \Cref{eq:attn-open-V}:
\begin{align}
     &\text{Attn}^{ \mC}( \vh)= \mW_v \vh \sigma\left( \vq ^T \mK\right)^T +  \vb_v\nonumber\\
     &\quad =\mW_v \vh\sigma\left( \vq ^T( \mW_k \mC+ \vb_k\mathbbm{1}^T)\right)^T  +  \vb_v \nonumber\\
     &\quad =\mW_v \vh\sigma\left( \vq ^T \mW_k \mC+ \vq ^T \vb_k\mathbbm{1}^T\right)^T  +  \vb_v . \label{eq:attn-open-2}     
\end{align}
Using the fact that $ \vb_k\mathbbm{1}^T$ is a matrix with identical columns, it follows that $ \vq^T \vb_k\mathbbm{1}^T$ is a vector with equal elements, where all elements are equal to $ \vq^T \vb_k$. On the other hand, softmax is invariant under translation by the same value in each coordinate. In other words, $\forall\delta \in \mathbb{R}, \forall \vz \in \mathbb{R}^n$, 
$\sigma( \vz+\text{\boldmath$\delta$}) =  \sigma( \vz)$, where $\text{\boldmath$\delta$}=\delta\mathbbm{1}$.
% \begin{align*}
%     \sigma( \vz+\text{\boldmath$\delta$}) =  \sigma( \vz) \quad\text{where}\quad \text{\boldmath$\delta$}=\delta\mathbbm{1}.
% \end{align*}
As a result, and from \Cref{eq:attn-open-2} we can conclude that,
\begin{align*}
     \text{Attn}^{ \mC}( \vh)= \mW_v \vh\sigma\left( \vq ^T \mW_k \mC\right)^T  +  \vb_v.     
\end{align*}
Next, in the equation above we replace $ \vq$ with its original linear transformation, resulting in,
\begin{align}
     \text{At}&\text{tn}^{ \mC}( \vh)= \nonumber \\
    &\mW_v \vh\sigma\left(( \mW_q \vh+ \vb_q) ^T \mW_k \mC\right)^T  +  \vb_v \label{eq:attn_rewritten}.     
\end{align}
This rewriting of attention highlights the different roles that $ \vb_q$, $ \vb_k$, and $ \vb_v$ play in the attention function. From \Cref{eq:attn_rewritten} it is clear that $ \vb_k$ plays \textbf{no role} in the attention function and it is in fact redundant. $ \vb_v$ on the other hand plays a very important role in attention since it is one of the two terms that are added to each other to constitute the attention function. Finally, $ \vb_q$ plays a role in creating the attention distribution along with other parameters of the attention namely, $ \mW_q$ and $ \mW_K$.

\section{Numerical Analysis}
\label{sec:comp}
In \Cref{sec:attn_dissec}, we mathematically show that the bias term of the \textit{key} linear transformation within the attention mechanism, i.e., $ \vb_k$, is redundant in the attention function and can be removed. In this section, we verify these results numerically. We also discuss some computational gains that could be achieved due to this result.

\begin{table*}[ht!]
  \small
  \centering
  \renewcommand{\arraystretch}{1.1}
  \begin{tabular}{|l||cccc||cccc||cccc|}
    \hline
     \multirow{2}{*}{}&
      \multicolumn{4}{c||}{$\vb_k$} &
      \multicolumn{4}{c||}{$\vb_q$} &
      \multicolumn{4}{c|}{$\vb_v$} \\
    \cline{2-13}
    & \textbf{0} & \textbf{1} & \textbf{10} & \textbf{[-5, 5]} &\textbf{0} & \textbf{1} & \textbf{10} & \textbf{[-5, 5]} & \textbf{0} & \textbf{1} & \textbf{10} & \textbf{[-5, 5]}\\
    \hline
    \hline
    \rule{0pt}{10pt}RoBERTa-base & $10^{-4}$ & $10^{-4}$ & $10^{-4}$ & $10^{-4}$ & $10$ & $10$ & $10^{2}$ & $10^{2}$ & $10$ & $10$ & $10^{2}$ & $10^{2}$ \\
    % \hline
    \rule{0pt}{10pt}RoBERTa-large & $10^{-5}$ & $10^{-5}$ & $10^{-5}$ & $10^{-5}$ & $1$ & $1$ & $1$ & $1$ & $1$ & $1$ & $10$ & $10$ \\
    % \hline
    \hline
    \rule{0pt}{10pt}BART-base & $10^{-5}$ & $10^{-5}$ & $10^{-5}$ & $10^{-5}$ & $10$ & $1$ & $10$ & $10$ & $1$ & $10$ & $10$ & $10$ \\
    % \hline
    \rule{0pt}{10pt}BART-large & $10^{-5}$ & $10^{-5}$ & $10^{-5}$ & $10^{-5}$ & $10$ & $1$ & $10$ & $10$ & $1$ & $10$ & $10$ & $10$ \\
    \hline
  \end{tabular}
  \caption{Tolerance level at which the final hidden state of the models before and after changing values of attention bias terms $\vb_k$, $\vb_q$, or $\vb_v$ for 100 sentences are equal. The elements of bias vectors are set to one of 0 (equivalent to removing), 1, 10, or a random value between -5 and 5. The models are not sensitive to even drastic changes to $\vb_k$.}
  \label{tbl:bias-change}
\end{table*}

\subsection{Changing \texorpdfstring{$ \vb_k$}{} in Pre-Trained Language Models}
\label{sec:change_b_k}
We examine the sensitivity of transformer-based pre-trained language models to changes to attention bias terms, i.e., $ \vb_k$, $ \vb_q$, and $ \vb_v$, for all attention modules within the model. The idea behind this analysis is that since we show that $ \vb_k$ is redundant in theory, changing its pre-trained values to arbitrary values should not impact the output of the models in practice. On the other hand, for $ \vb_q$ and $ \vb_v$, which are not redundant in the attention function, changing their values should result in significant changes in model outputs.

For this analysis, we take the first 100 sentences from the Wikipedia page for ``Machine Learning''\footnote{https://en.wikipedia.org/wiki/Machine_learning}. The length of these sentences ranges from 1 (``Overview'') to 75 tokens (counted using spaCy\footnote{\texttt{en_core_web_sm} package. https://spacy.io/}). 
Next, we feed these sentences to several pre-trained language models and get their last layer's hidden state for each sentence. We represent this hidden state matrix for sentence $i$ as $\mH_i \in \mathbb{R}^{d \times s_i}$, where $s_i$ is the length of sentence $i$.
We also feed these sentences to the same pre-trained language models but with changes applied to their $ \vb_k$, $\vb_q$, or $ \vb_v$ (details are forthcoming) and represent the corresponding hidden state matrix as $\mH'_i$, which is also in $\mathbb{R}^{d \times s_i}$. 
We then compare $\mH_i$ and $\mH'_i$ across all $i$s for each of the models: 

\begin{equation*}
\label{eq:tol}
x^{\star} := \inf\left\{ x\in\mathbb{Z} \mid \max_{i} \| \mH_i - \mH'_i \|_{\text{max}} \le 10^x \right\},
\end{equation*}

where $\| . \|_{\text{max}}$ is matrix max norm
\footnote{https://en.wikipedia.org/wiki/Matrix_norm\#Max_norm}, $\mathbb{Z}$ is the set of all integers, and $i\in\{1, \dots, 100\}$.
In other words, the \textit{tolerance level} (i.e., $10^{x^{\star}}$) at which $\mH_i$ and $\mH'_i$ are equal across all sentences is calculated.

We run this analysis for base and large sizes of both RoBERTa \cite{Liu2019RoBERTaAR} and BART \cite{lewis-etal-2020-bart} using Huggingface.
%\cite{https://doi.org/10.48550/arxiv.1910.03771}. 
We set the values of the elements of bias vectors $\vb_k$, $\vb_q$, or $\vb_v$ to 0, 1, 10, and random numbers uniformly sampled from $[-5, 5]$. In other words, $\vb_k$, $\vb_q$, or $\vb_v$ is set to a vector of zeros, ones, tens, or a random vector. This is done for all attention modules (e.g., both self- and cross-attentions in BART) within the models. A small python script for this is shown in the Appendix. The tolerance levels ($10^{x^{\star}}$) for this analysis are reported in Table \ref{tbl:bias-change}.
For both BART-base and -large, as well as for RoBERTa-large, we see the models are not sensitive to the values of $\vb_k$ at $10^{-5}$ tolerance level. For RoBERTa-base this tolerance level is $10^{-4}$. On the other hand, for the other two attention bias terms $\vb_q$ and $\vb_v$, the models are very sensitive to changes to these bias terms. For instance, for RoBERTa-base, changing the elements of $\vb_q$ or $\vb_v$ to random values in $[-5,5]$, results in the elements of the final hidden states to change up to 100 in value. This study shows that $\vb_k$ does not have any significant impact on the output of the models.

From the conclusion of \Cref{sec:attn_dissec} that $\vb_k$ is redundant in the computations of attention, one might expect that the tolerance levels reported in Table \ref{tbl:bias-change} under $\vb_k$ should be much lower. This discrepancy is simply due to numerical errors associated with calculating softmax within the attention function. For example, in theory, softmax of $[0.1,0.2]$ is equal to the softmax of $[5.1, 5.2]$, since softmax is invariant under translation by the same value in each coordinate. However, numerically this equality only holds at $10^{-9}$ tolerance level\footnote{Calculated using the implementation of softmax in the nn module of PyTorch version 1.10.0}. 
These errors propagated across hundreds of dimensions (instead of 2 in this example), and numerous transformer layers would lead to tolerance levels of $10^{-4}$ and $10^{-5}$ that are reported in \Cref{tbl:bias-change}.

We conduct a similar experiment on text classification task and measure how changes in $\vb_k$, $\vb_q$, and $\vb_v$ impact the accuracy of models. For this purpose for three GLUE \cite{DBLP:journals/corr/abs-1804-07461} tasks, namely SST-2, MNLI, and QQP, we take pre-trained RoBERTa-large based models and change the values of $\vb_k$, $\vb_q$; and $\vb_v$ to random values uniformly sampled from $[-5, 5]$; we report accuracy of models on the validation set before and after these changes in \Cref{tbl:glue}. From the numbers, it is immediately clear that applying these drastic changes to $\vb_k$ results in no change in the accuracy of the models. On the other hand these changes to $\vb_q$ and $\vb_v$ result in very large degradation in the performance of the models. It is also evident that changes in $\vb_v$ result in much larger degradation than changes in $\vb_q$, which supports the evidence in \Cref{sec:attn_dissec} about role of $\vb_v$.

\begin{table}[t!]
  \footnotesize
  \centering
  \begin{tabular}{|l||c|c|c|c|}
  \hline
  & & \multicolumn{3}{c|}{\textbf{[-5,5]}} \\
  \hline
   & Original & $\vb_k$ &	$\vb_q$ &	$\vb_v$ \\
  \hline
    SST-2\tablefootnote{https://huggingface.co/philschmid/roberta-large-sst2} & 0.9644 & 0.9644 &	0.7007 & 0.4908 \\
    MNLI\tablefootnote{https://huggingface.co/roberta-large-mnli} & 0.9060  & 0.9060 & 0.4101 & 0.3182 \\
    QQP\tablefootnote{https://huggingface.co/howey/roberta-large-qqp}	& 0.9214 & 0.9214 & 0.6434 & 0.3682 \\
  \hline

  \end{tabular}
  \caption{Accuracy of GLUE \cite{DBLP:journals/corr/abs-1804-07461} tasks as the values of $\vb_k$, $\vb_q$, and $\vb_v$ of  trained models, which are based on RoBERTa-Large, are set to uniform random values between -5 and 5.
}
  \label{tbl:glue}
\end{table}

\subsection{Pre-Training of Language Models without \texorpdfstring{$ \vb_k$}{}}
We train two transformer-based language models, GPT-2 \cite{radford2019language} and RoBERTa, from scratch both with and without $ \vb_k$, and compare the language modeling performance of the model on a test set. We use Huggingface with the original hyper-parameters for training these models. Both of these models are trained on wikitext-103-v1 \cite{DBLP:journals/corr/MerityXBS16} dataset. We train each of the GPT-2 (small) and RoBERTa (base) models from scratch with three different random seeds for 50,000 steps, and we report the loss of final model on the test set averaged over three runs in Table \ref{tbl:lm}. Note that there is no statistically significant difference between the two settings at a p-value $\leq$ 0.05 in an unpaired two-tailed T-test.

\begin{table}[t!]
    \small
    \centering
    \renewcommand{\arraystretch}{1.5}
        \begin{tabular}{|l|c|c|}
        % \toprule
            \hline
            & \textbf{GPT-2} & \textbf{RoBERTa-base}  \\
            \hline
            Original & 2.9251 &	5.8890 \\
            \hline
            No $ \vb_k$ & 2.9250	& 5.8909 \\
            \hline
        % \bottomrule
        \end{tabular}
    \caption{Average loss (on test set) of GPT-2 (small) and RoBERTa-base compared to their variants without $ \vb_k$,  trained from scratch. Numbers are average over 3 runs with different random seed, and no statistically significant difference is observed.}
    \label{tbl:lm}
\end{table}

\subsection{BitFit without \texorpdfstring{$ \vb_k$}{}}
\label{sec:bitfit}
One place where removing the redundant $ \vb_k$ could result in significant savings in computation is in BitFit \cite{ben-zaken-etal-2022-bitfit}, where a pre-trained transformer based language model is fine-tuned for downstream tasks such as text classification, summarization, etc., by freezing all the trainable parameters of the model, except for bias terms  within different modules of the model. This is further discussed in details in \Cref{sec:bitfit}.

Next, we study the effect of freezing $\vb_k$ vectors across all transformer layers in BitFit. Normally in BitFit, $\vb_k$ vectors are among the fine-tuned parameters, but since we show that $\vb_k$ is redundant in the attention function, we study what happens if these vectors are not fine-tuned in BitFit. In this section all models are fine-tuned using the exact hyper-parameters used in \cite{DBLP:journals/corr/abs-2110-04366} for the corresponding models. 
Table \ref{tbl:xsum} shows the results of BitFit for summarization task on the XSUM \cite{narayan-etal-2018-dont} dataset, using BART-large with and without fine-tuning $\vb_k$. Freezing $\vb_k$ in BitFit results in 11.1\% decrease in the number of trainable parameters. Each model is fine-tuned three times with different random seeds. The reported numbers in \Cref{tbl:xsum} are different rouge metrics averaged over the three runs. According to a two-tailed T-test there is no statistically significant difference between the rouge metrics at p-value $\leq$ 0.05.

\begin{table}[t!]
  \footnotesize
  \centering
  \begin{tabular}{|l|l|l|l|l|l|}
  \hline
  & & \multicolumn{4}{c|}{\textbf{Average Rouge}} \\
  \hline
  & & R-1 &	R-2 &	R-L &	R-Sum \\
  \hline
  \multirow{2}{*}{BART-large} & Orig. & 40.66 &	17.32 &	32.25 &	32.25 \\
  & No $\vb_k$ & 40.67 & 17.33 &	32.26 &	32.26 \\
  \hline

  \end{tabular}
  \caption{BitFit on BART-large with/without $\vb_k$ on XSUM. Second row has 11.11\% less trainable parameters than the first row. 
  No statistically significant difference in the evaluation set accuracy according to a two-tailed T-test at p-value 0.05.
}
  \label{tbl:xsum}
\end{table}

We conduct a similar experiment for a text classification problem, namely SST-2 from the GLUE benchmark using RoBERTa. The main difference between this and the summarization setting is the new classification layers that need to be added to RoBERTa for performing classification, whose parameters are fine-tuned along with the bias parameters in BitFit. As a result, the savings in the number of trainable parameters by freezing $\vb_k$ in this setting is smaller than the summarization setting with BART. For RoBERTa-base and RoBERTa-large freezing $\vb_k$ during fine-tuning using BitFit results in 1.3\% and 1.9\% savings in trainable parameters, respectively.
Table \ref{tbl:sst} shows the average accuracy (over five runs with different random seeds) of the fine-tuned models on the evaluation set for SST-2. According to a two-tailed T-test, at p-value $\leq$ 0.05 there is no statistically significant difference between the BitFit results with and without $\vb_k$ for both base and large sizes of RoBERTa.

\begin{table}[ht!]
  \footnotesize
  \centering
  \begin{tabular}{|l|l|c|}
  \hline
  & & \textbf{Eval. Accuracy} \\
  \hline
  \multirow{2}{*}{RoBERTa-base} & Original & 94.98 \\
  & No $\vb_k$ & 94.92  \\
  \hline
  \multirow{2}{*}{RoBERTa-large} & Original & 95.83 \\
  & No $\vb_k$ & 95.85  \\
  \hline

  \end{tabular}
  \caption{Average accuracy over 5 runs of BitFit on RoBERTa-base and -large with and without $\vb_k$ on SST-2. No statistically significant difference in accuracy is observed according to a two-tailed T-test at p-value 0.05.}
  \label{tbl:sst}
\end{table}

\section{Implications for Transformers}
% \label{sec:implications}
As was shown in the previous sections, the bias term of the \textit{key} linear transformation, i.e., $ \vb_k$ in the attention function is redundant. In the context of transformers, if we consider one transformer layer, $ \vb_k$ constitutes only less than 0.01\% of the parameters of the layer. As a result, removing $ \vb_k$ from the transformer architecture both during training or even from a pre-trained model at inference time does not result in significant savings in computations.
% As was discussed earlier, the bias term of the \textit{key} linear transformation in attention mechanism that is widely used in transformers constitute a small fraction of all the parameters of the transformer models. 
However, for the following reasons we argue that this finding is important. (1) The small size of these redundant parameters within one of the most widely used neural networks architectures does not change the fact that they are redundant, and we argue that this redundancy, however small, should be addressed. (2) It is important to note that this small redundancy appears in thousands of transformer-based models that are being invoked millions of times a day across academia and different industries within different products. From this aggregate perspective this redundancy results in significant redundant computational operations that could be avoided. (3) Recent works such as $(IA)^3$ \cite{https://doi.org/10.48550/arxiv.2205.05638} show how a small set of parameters (of the same size as $b_k$) could be used for efficiently adapting large language models to downstream tasks (often rivaling fully fine-tuned variants). From this angle, the redundant $b_k$ parameters could be repurposed to improve the performance of models. (4) Finally, in some recent works the bias terms of ``dense kernels'' and ``layer norms'' are dropped from the architecture based on the observation of positive impact on stability of the training process. Our analysis reveals that from a theoretical standpoint some additional biases ($b_k$) could be dropped from these architectures as well.

\section{Conclusions}
\label{sec:conc}
In this work, we analyze the attention function predominant in present-day transformer architectures and find that 
the biases in the linear transformations of the attention function play different roles. Our analysis reveals that $\vb_v$ is an important component in the attention computation, whereas 
the bias of the key linear transformation, $\vb_k$, is completely redundant. We also numerically confirm that removing $\vb_k$ does not significantly change the outcome of transformers.

While our analysis has been focused on the softmax-based (scaled) dot-product attention, recent works have demonstrated how attention can be generalized from the kernel lens. This has led to innovations in different kernel designs that perform equivalently or better
%~\cite{likhomanenko2021cape,qin2021cosformer}, 
and it would be interesting to explore the role of bias terms in the proposed kernels.

\label{sec:refs}
\bibliographystyle{IEEEbib}
% \bibliography{strings,refs}
\bibliography{main}

\end{document}

%% file: math_commands.tex
\usepackage{amsmath,amsfonts,bm}

\def\eqref#1{equation~\ref{#1}}
\def\1{\bm{1}}

\def\vb{{\bm{b}}}

\def\vh{{\bm{h}}}

\def\vq{{\bm{q}}}

\def\vz{{\bm{z}}}

\def\mC{{\bm{C}}}

\def\mH{{\bm{H}}}

\def\mK{{\bm{K}}}

\def\mV{{\bm{V}}}
\def\mW{{\bm{W}}}

\DeclareMathAlphabet{\mathsfit}{\encodingdefault}{\sfdefault}{m}{sl}
\SetMathAlphabet{\mathsfit}{bold}{\encodingdefault}{\sfdefault}{bx}{n}